# Optical flow GNSS for navigation in the Indian subcontinent (NavIC)

Sunit Shantanu Digamber Fulari

Department of Electronics and Communication
Chandigarh University, Mohali, India

**Abstract.** This paper reveals about global navigation satellite system GNSS in the indian subcontinent known as the navigation in the indian subcontinent(NavIC) We have tried to model a new technique in GNSS known as the optical flow tracking global navigation system (OF GNSS). This method using differential equations is very accurate for very small distances on the surface of the earth in the 1500km range of the Indian subcontinent satellite coverage. When we talk of accuracy of the GPS system it should be very accurate on the surface of the earth when used to show changes in coordinate of the moving body with respect to the ground by the satellite which is situated on the earths orbit. Optical flow is a method which uses movements with respect to x and y axis for infinitesimal changes in its coordinates and then uses this algorithm to use it in global positioning system to find accurate position of the body with respect to the satellite coordinates with respect to ground positioning. The modern method of differential frames is also very accurate as it involves infinitesimal frames which are modelled together from the satellite to find changes in the coordinates on the earths surface, so we have designed a new algorithm in this paper on the Optical flow GNSS system which is an alternative and can improve the study done in the design of these algorithms in this field of applications.

## 1 Introduction

- The indian subcontinent coverage is more than 1500km known in satellite terms as NavIC or navigation in the indian subcontinent.
- We have modelled a optical flow GNSS (OF GNSS) which is accurate to very small distances.

- We have also tried to provide a summary on the various GNSS radiation methods like VT DFVT. The others like VT DFVT being applied to modern and latest techniques have been found to be more accurate than other older techniques using L5 band which were found to find these defects or drawbacks in its results and applications such as obstructions due to scintillation effects, ionospheric effects, tropospheric effect, interference, multipath effect and other major readings which could not have been possible due to obstructions by buildings and other ground surface bodies.
- Our method is derived for small differences in space detected by global positioning system(GPS) which identifies movements of a body on the ground surface with coordinates axis which can identify motion of the body. This motion is nothing but positioning of the body on the satellite axis corordinate system. The is shown linearity of this system with other pre originated methods which show similar effect and accuracy, the other being researched earlier.
- Why is our method (OF GNSS) is more accurate than other traditional methods which involved more complex derivations and methods which were otherwise not required in this. These algorithms of Optical flow GNSS is maybe underlooked and not studied before as it involves the fundamental theorem of calculus in its domain, they were not studied before, they have been involved of the basic study, the fundamental theorems, proving to be very effective in determining the position of a body on the coordinate system.
- The mathematical interpretation of this method is but the equation of a line.

The optical flow equation is derived for small differences in space derived using differential equation. In the first step we use finite infinitesimal differences in space, find partial differential equation of it, followed by further fourier transform to obtain the final equation which is equation of a line.
This optical flow is used in satellite navigation system which is a system derived by our proposed method.

## 2 Review of previous papers

### 2.1 Carrier aided dual frequency vectorized tracking architecture for NavIC signals

This paper uses accuracy improvement known as vector tracking in indian subcontinent(NavIC) using dual frequency vectorized tracing. It utilizes s band and L5 noise cancellation band to mitigate and use extended kalman filter to detect signals NavIC15 and CA DFVT with respect to time, position and velocity. This method to improve accuracy minimizes ionospheric losses and other which were visible in the previous methods. The previous system of decentralized ST system in GNSS was very inaccurate in various extreme situations. Dual frequency vector tracking is easy installation and low cost with ease of implementation. It is used in extreme conditions of non line of sight, low signals , extreme turbulence, interference and other conditions leading to issues.
The availability of GNSS has made such possibility of time and position measurements of a body in wide application environments such as rail, civil signals, aviation, surveying, agriculture, timing and in providing location services to users. Today there is only one Global positioning system available to all the users and if that is attacked or hacked then there is no alternative available, so we have to find a secondary subsidiary which will function in case our primary working GPS system fails. Among those threats is GNSS spoofing. In the paper [7] they have very easily shown kalman based spoofing detection method

The following section deals with a small study on the antenna used for terrestrial communication. We have tried to simulate a horn shaped antenna, we have found some results which are written formed in this paper. horn shaped antenna is used for terrestrial satellite communication, we formulate such type of antenna which can be used for satellite to outer space and earth observatory communication. We figure out a link when this horn shaped antenna communicates by sending signals to other linked receivers. When this horn shaped antenna with very high efficiency is used, we have written this type of similar horn shaped antenna study in our previous papers briefly in this antenna design(horn shaped). This is also known as microwave antenna with flaring structure protruding at its end.
Figure II:Far field at theta 90 degrees
This design of far field in planar polarization of electromagnetic wave varies from 0 to 14 degree decibels and obtains a shape in the form of embedded human teeth kind of diagram.

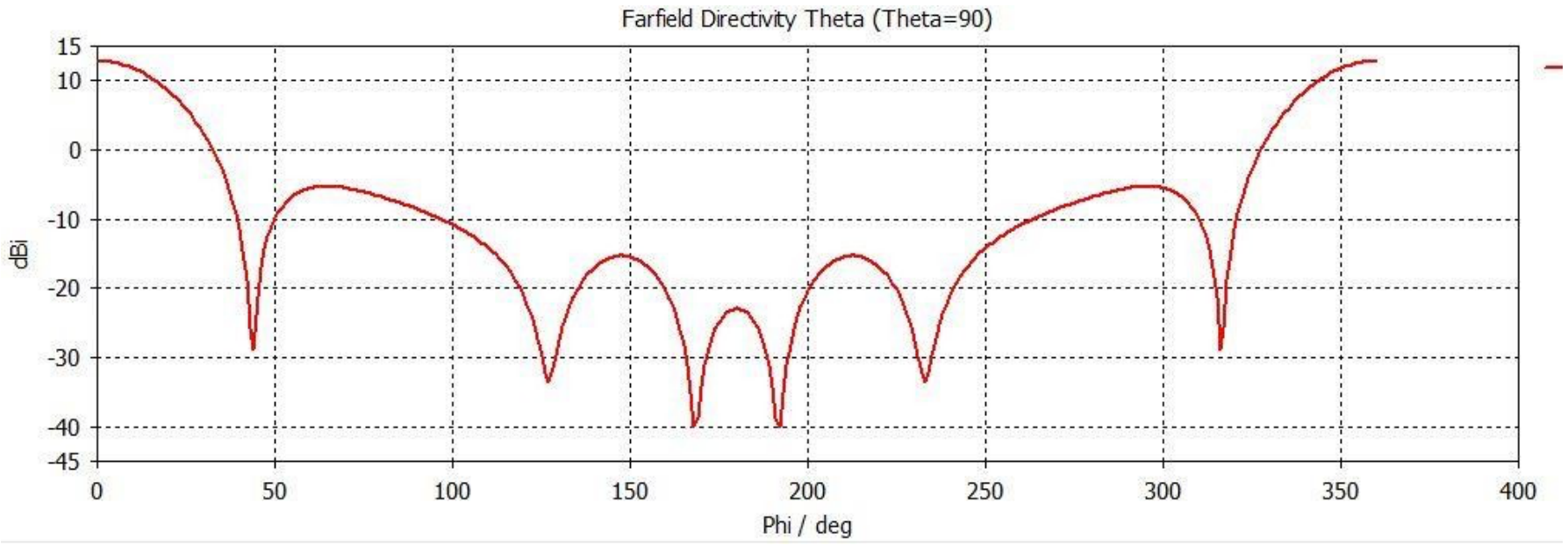

Frequency = 39 GHz
Main lobe magnitude = 12.8 dBi
Main lobe direction = 0.0 deg.
Angular width (3 dB) = 33.6 deg.
Side lobe level = -18.1 dB

Far field directivity is also an important criteria to study in this form of so called horn shaped antenna. In the case of far field region in play the electromagnetic wave attains to that state when there is polarization of the outgoing wave independent of the distance from the antenna and receiver antenna. While the close or near field is analysis at distances very close from the antenna and electromagnetic radiation when the distance from the sender to the receiver is very small which is rare and takes place in very close mobile phone communication separated by distances of a few metres. We have highlighted far field pattern which is widely used and overall communication mainly occurs in far field region.

Figure VII:far field directivity at 28,33.5 and 39 Ghz for far field at phi=0 degrees

The figure shows a improved far field in terms of 28Ghz which decreases from 33.5Ghz to 39Ghz respectively. 28Ghz shows linearlity while 39Ghz shows unmarginal variability and descent and large gradient. The figure shows a constant value of frequency waves subsidiary at frequency=28. 33.5 and 39 frequency shows curves going into a loop, known in the millimetre wave antenna range.

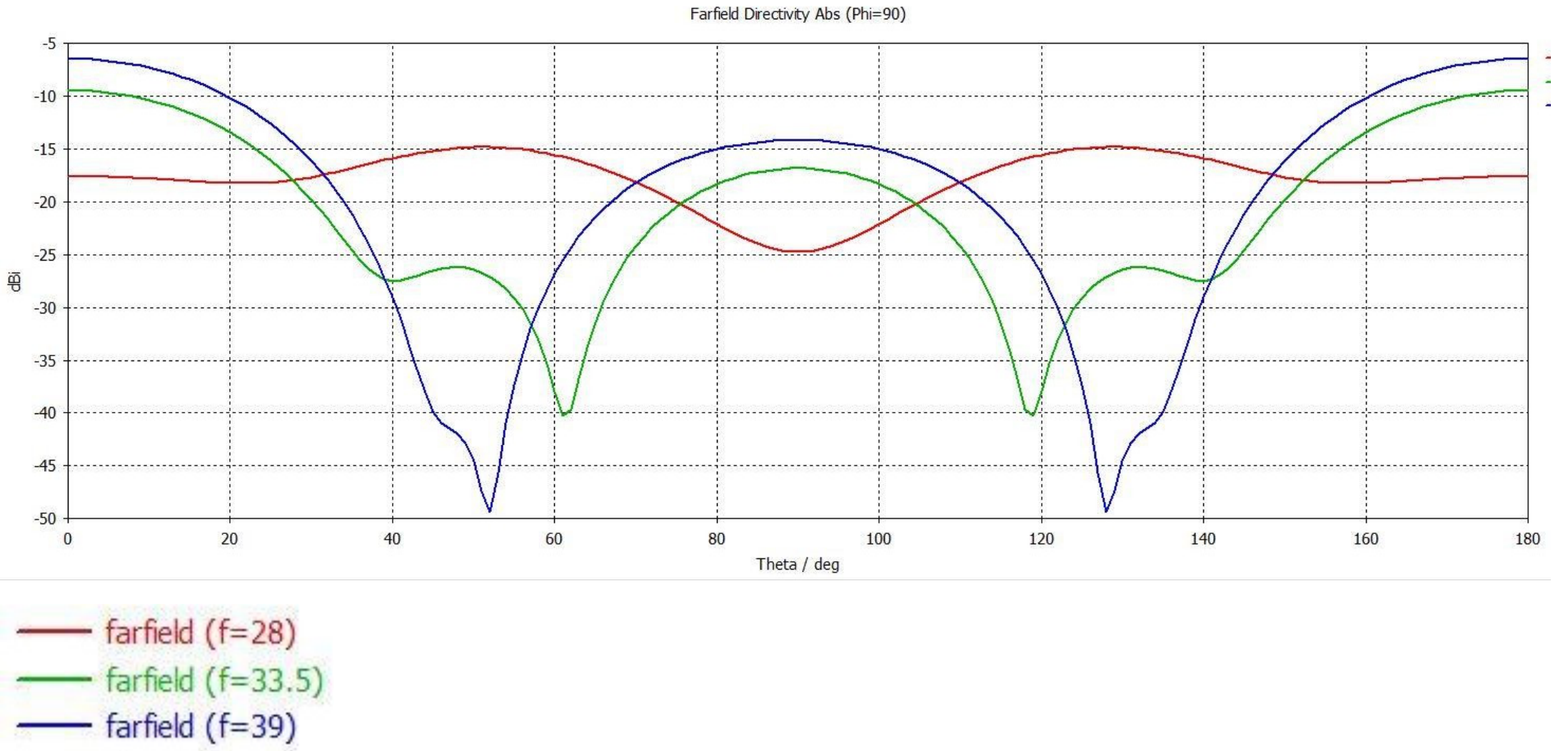

All the three freqeencies 28,33.5 and 39 Ghz show uniform variation and marginal changes in this case of theta at 90 degrees.

The optical flow equation is derived using the standard equation

$$I(x_{im}, y_{im}, t)$$

X and y are positions on the coordinate for interval time $I(x_{im}, y_{im}, t)$=$I(x_{im} + \partial x, y_{im} + \partial y, t + \partial t)$

Using taylor series we obtain

$$I_x v_{xim} + I_y V_{yim} + I_t = 0$$

X and Y are components of optical flow.

Our proposal is better in terms of accuracy and the theory is augmented in motion detection. The optical flow equation opens wide scope for object detection in GPS system of GNSS better than the previous methods proposed. The object is detected for small differences in 3D space on the ground for small metres distances.

There is accuracy to the measure of metres in this case when optical flow is combined in s band detection.

The gps tracker plot is given in the figure below
Figure X:

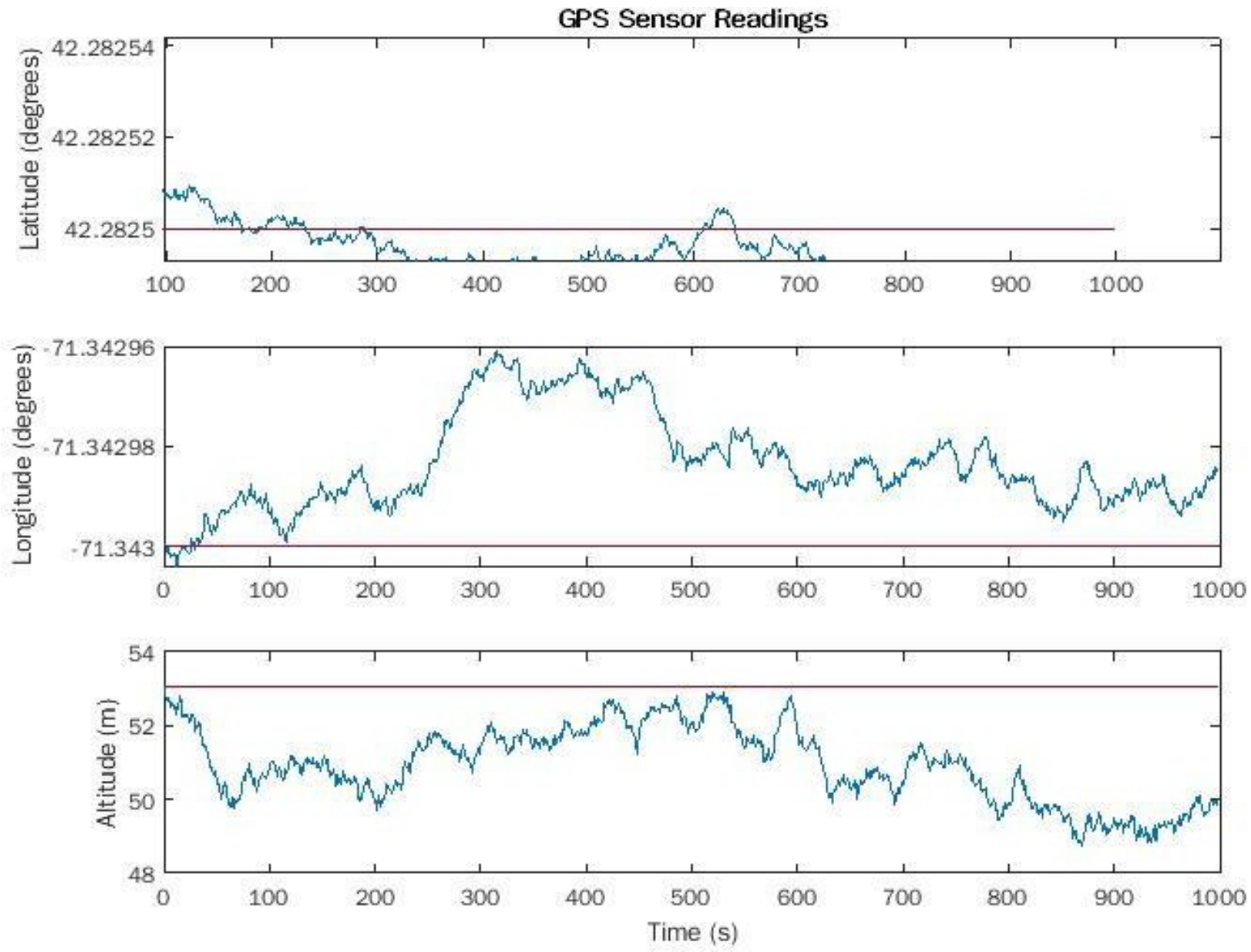

The altitute about latitude and longitude is given by the equation and plot below. There is seen a major deviations in detection on the GNSS system. The figure above is of GPS sensor readings. There is longitude variation in the prime meridian from 42 degres from time 100 to 720 seconds time. This position, time and velocity accuracy is achievable in our design. The second sub diagram shows shows high variability at the median while the third sub diagram shows uniform variability which shows the efficiency and stability of the device. Optical flow GNSS (OF GNSS) shows greater variability in terms of the variations in longitude and prime meridian. The efficiency of the device is 10-20% higher than the other traditional methods. When we use this deivce in positions and situations of high interference it is also very viable and stable which shows high accuracy in such environments. The variations is from 0 to 1.667 minutes.

Figure XI:

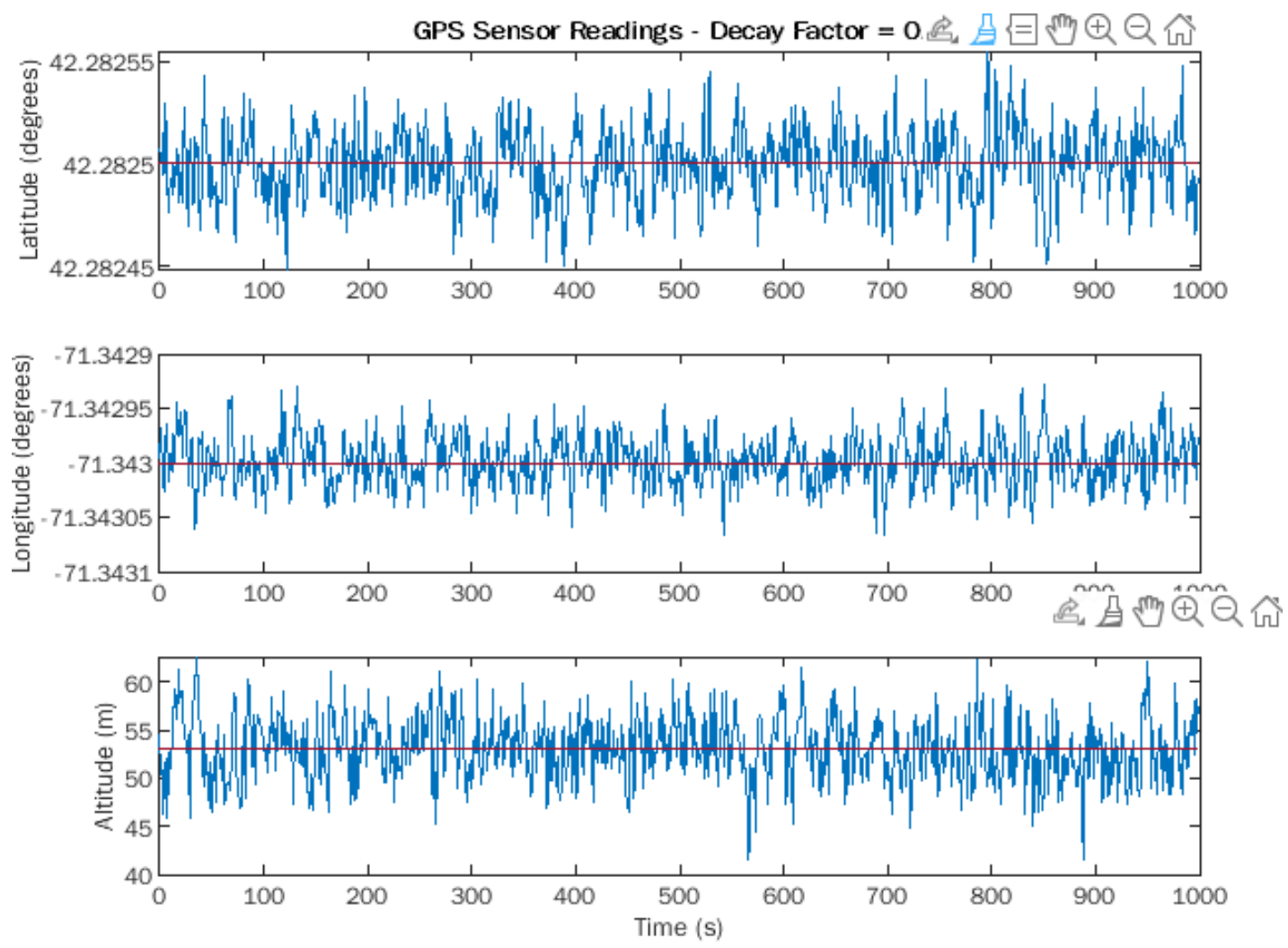

The GPS sensor reading vary to a greater extent from 42.28245 to 42.28255 in the first, from -71.34305 to -71.34295 and from 40 to 63 in the third. These variations depict that though there are slight variations in the ground GPS which can be improved in the future algorithms. 0-4 metres difference in the position is improved in this case.

Figure XI: When a GPS device is placed on the ground surface, the measured accuracy with respect to the ground GNSS is given in thefigure above.

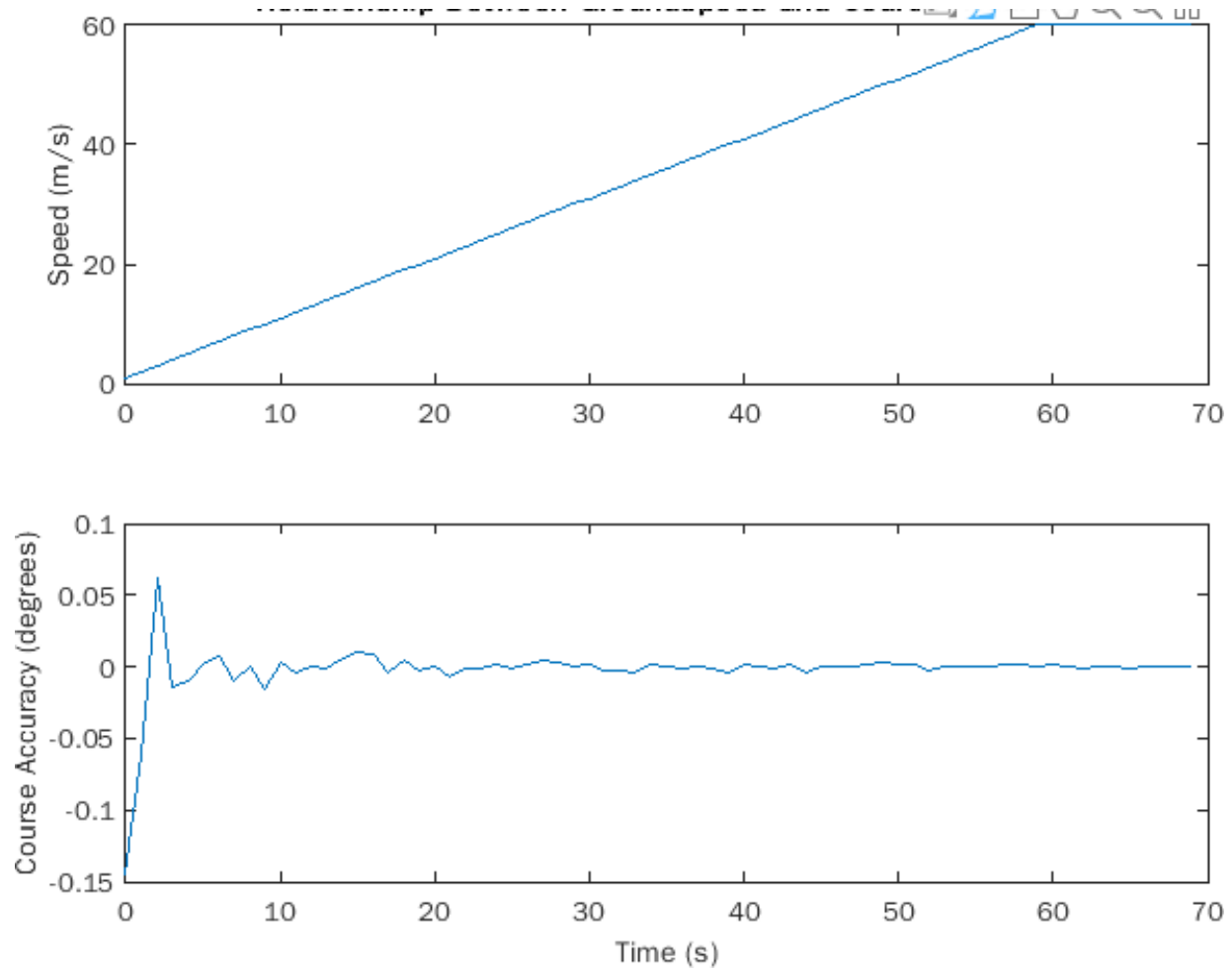

The optical flow for GNSS system will detect objects in the frequency range from 28-39Ghz with minimum latency. The observed theorem has high efficiency.
Figure XII: Location identifier

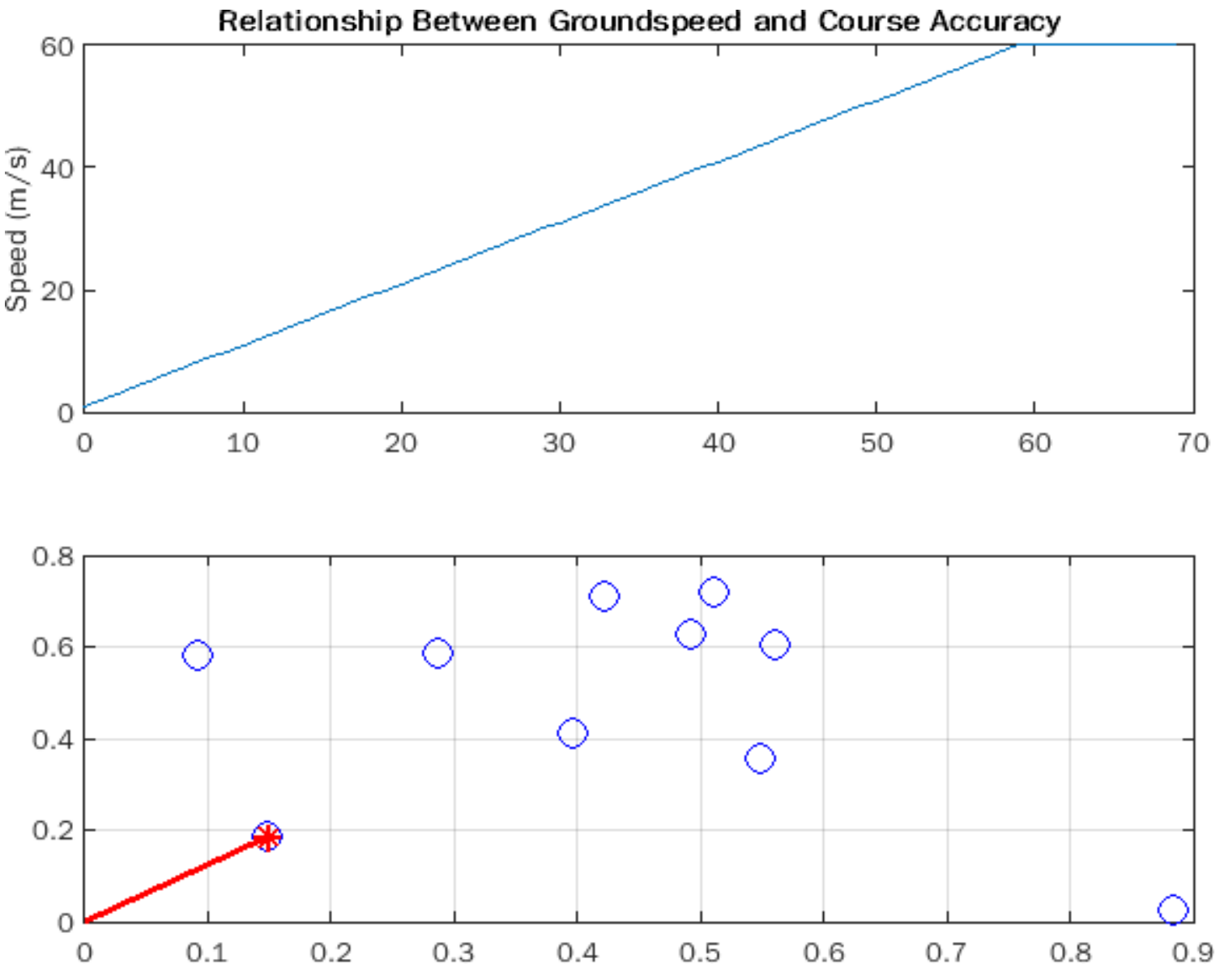

In our Optical flow OF GNSS when there is connectivity between the movement of the device on the ground surface the simulation is achieved as above in figure XII. The figure XIII is another diagram showing the efficiency of the device connecting between small distances on the surface. The points are matched to find the distance on the surface of the earth using the satellite, the closest distance is found for the moving vehicle, the points shows the matching conducted experiment to find for the motion using optical flow equation.
Figure XIII:

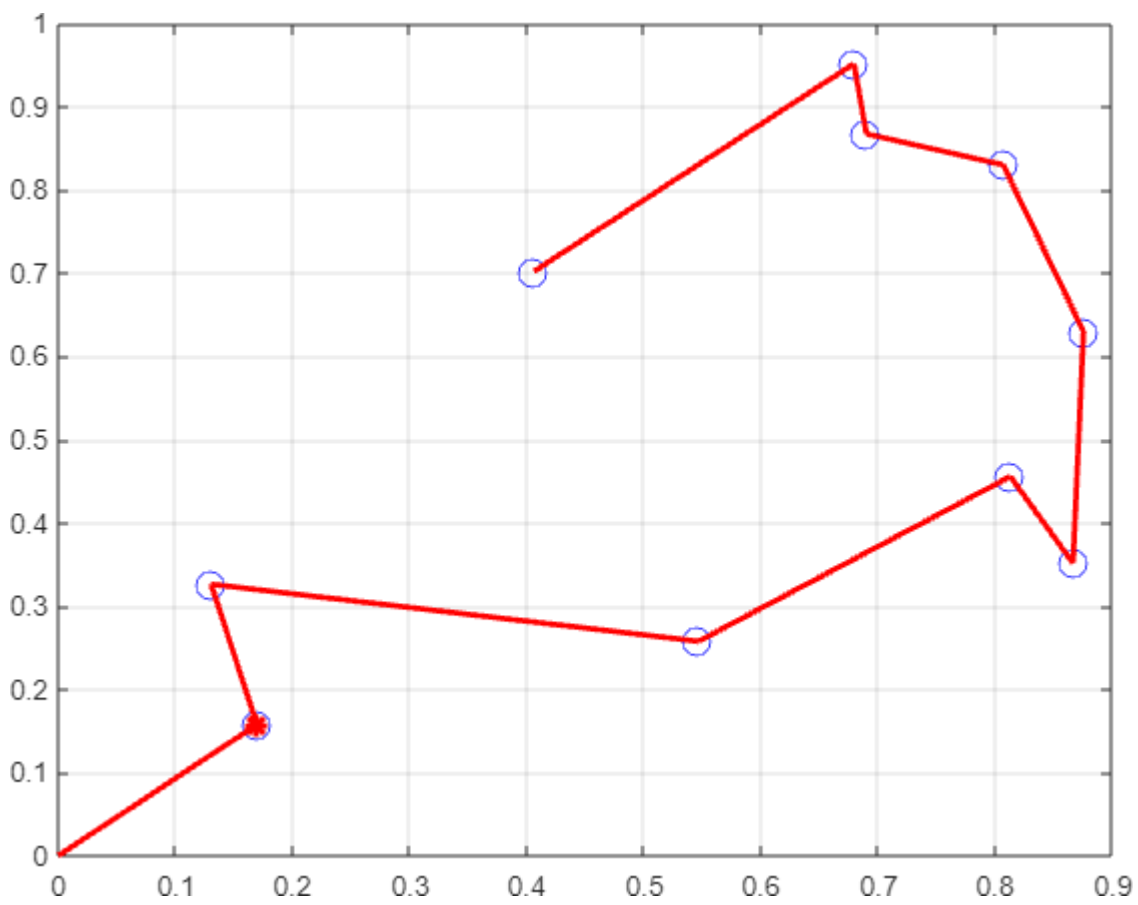

How the connection between different devices in towers takes place in Global positioning system.

The extended kalman filter is used in detecting the motion of bodies, in this case the aircraft which is moving in the sky waves. Estimated position of the optical flow GNSS is shown to be with respect to the actual position in the apparatus shown below. There is a visualization of a aircraft motion and sensor which is used to make the aircraft move in a defined path in the space, this is done with a sensor in the aircraft which is handled by the pilot in the cockpit, used throughout takeoff, path movement and landing in the airport.

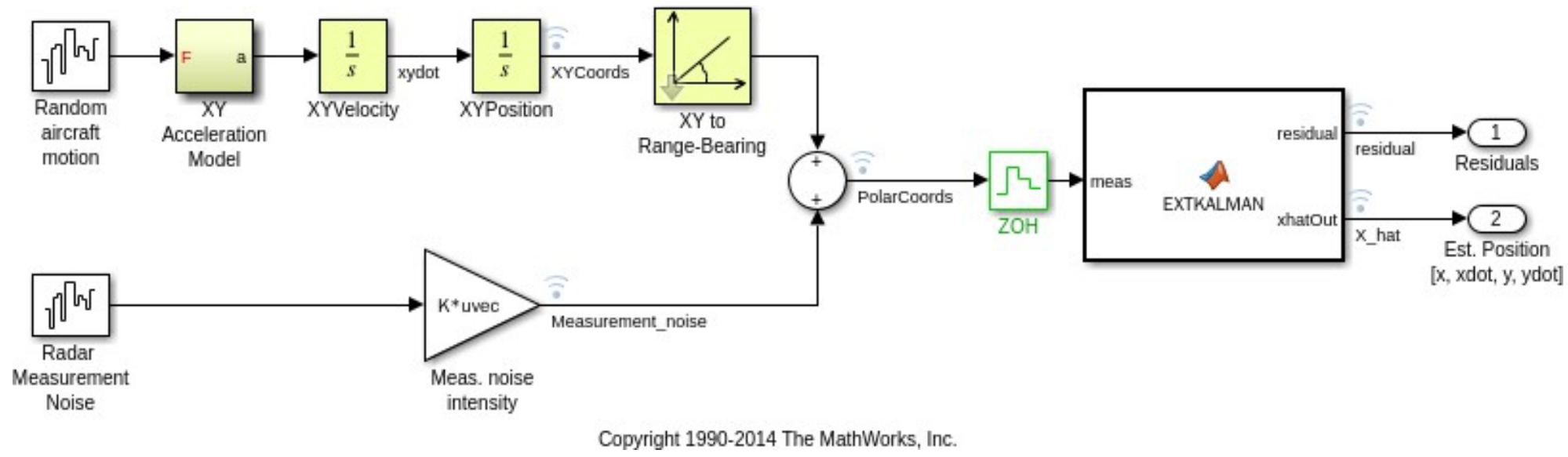

Figure XIV:Waves generated

The above combination of random aircraft connected with acceleration model linked to velocity and position to distance measurement. There is depiction of range(position), velocity and time of the device. These parameters are very important in our study. Noise or error is minimized.

There was a time about half a century ago when there was no such technology available to us, and there was extreme difficulty in path detection and map traversing. There was no global positioning system thought about that time. Today with the introduction of this system a traveller can easily install the system on his device(mobile device, GPS system or any other gadget) this is used to detect motion on the surface of the earth of the body, this is connected to the satellite system, there is no limit to how many devices use this system on the earth system from the satellite. There are today all the android and ios apps available such as strava app which is used in route history in terms of the workouts used in the system. This system has made an overall new world in terms of global positioning sytem.

Figure XV:Shows a GPS sytem of Strava app on workout issues using a different algorithm , they charge about 7 dollars per month, which is the only system available to all users, if this system gets hacked or destroyed in space attacks then there is no alternative. Our algorithm can be used in the future if it can be found viable in this applications.

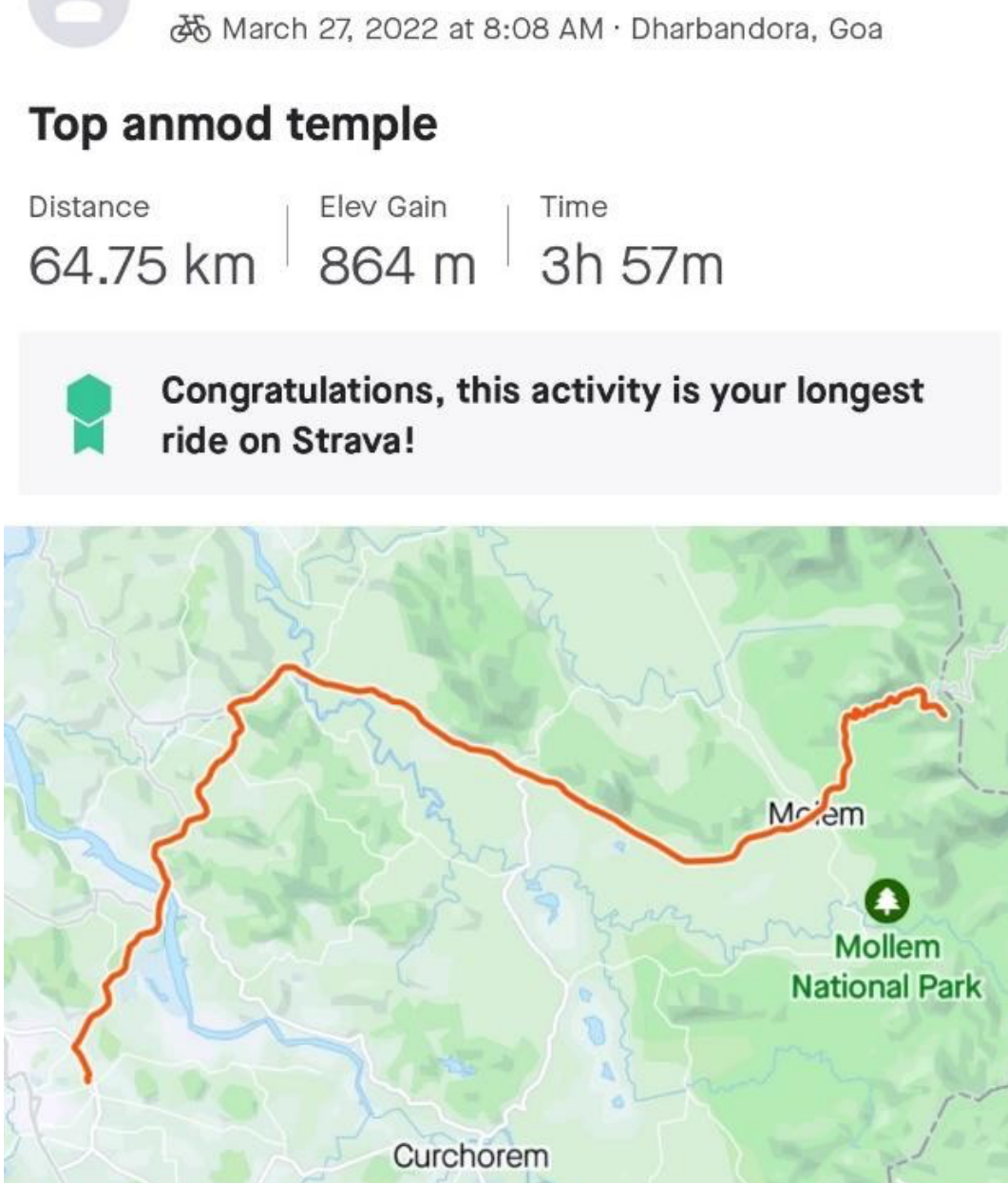

This system is being used in all the applications which display route to its users ranging from vehicle route tracking to any of the application similar to the above figure. We have in our paper shown a different algorithm which can be implemented in such applications in a different way(such that in future satellites which may find its applications). The path is correctly shown even in places where there is no internet access or connectivity. This is an added advantage as there is no mobile data(internet) wasted during the path or in remote areas where there is no availability. Only required during uploading on the server or app. We want to adopt and implement OF GNSS in this app similar to strava app to be used in tracking in the indian subcontinet(TrackingIC).

Figure XVI:The optical flow GNSS detection in blue line varies with the actual yellow line.

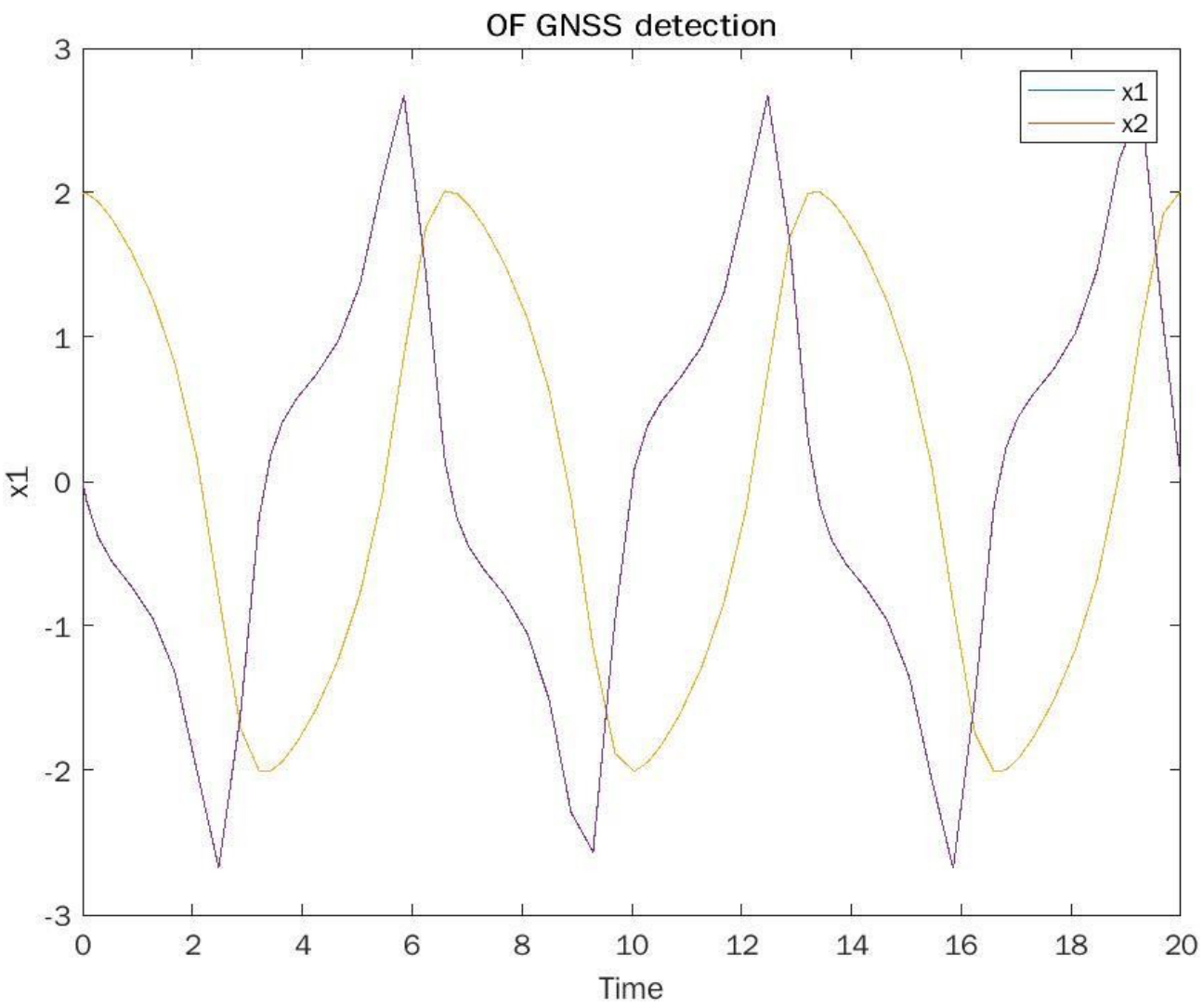

When the object is accelerating with acceleration on the ground the plot in GNSS is given as below on the surface(body). This motion is shown as a waves in terms of our proposal on OF GNSS detection.

Figure XVI:Showing acceleration of the device on the ground place with the changes in time. There is uniform motion and changes seen in the figure. Sometimes accuracy is sacrificed in GPS system detection, this happens when there is large motion of body on the ground surface, and there is small or large error in tracking

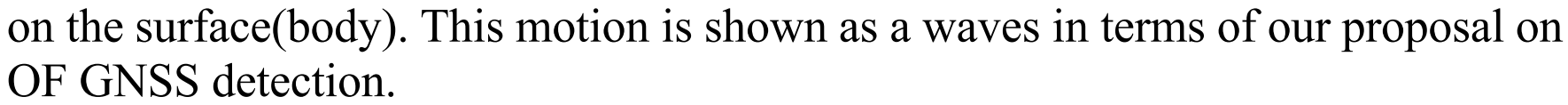

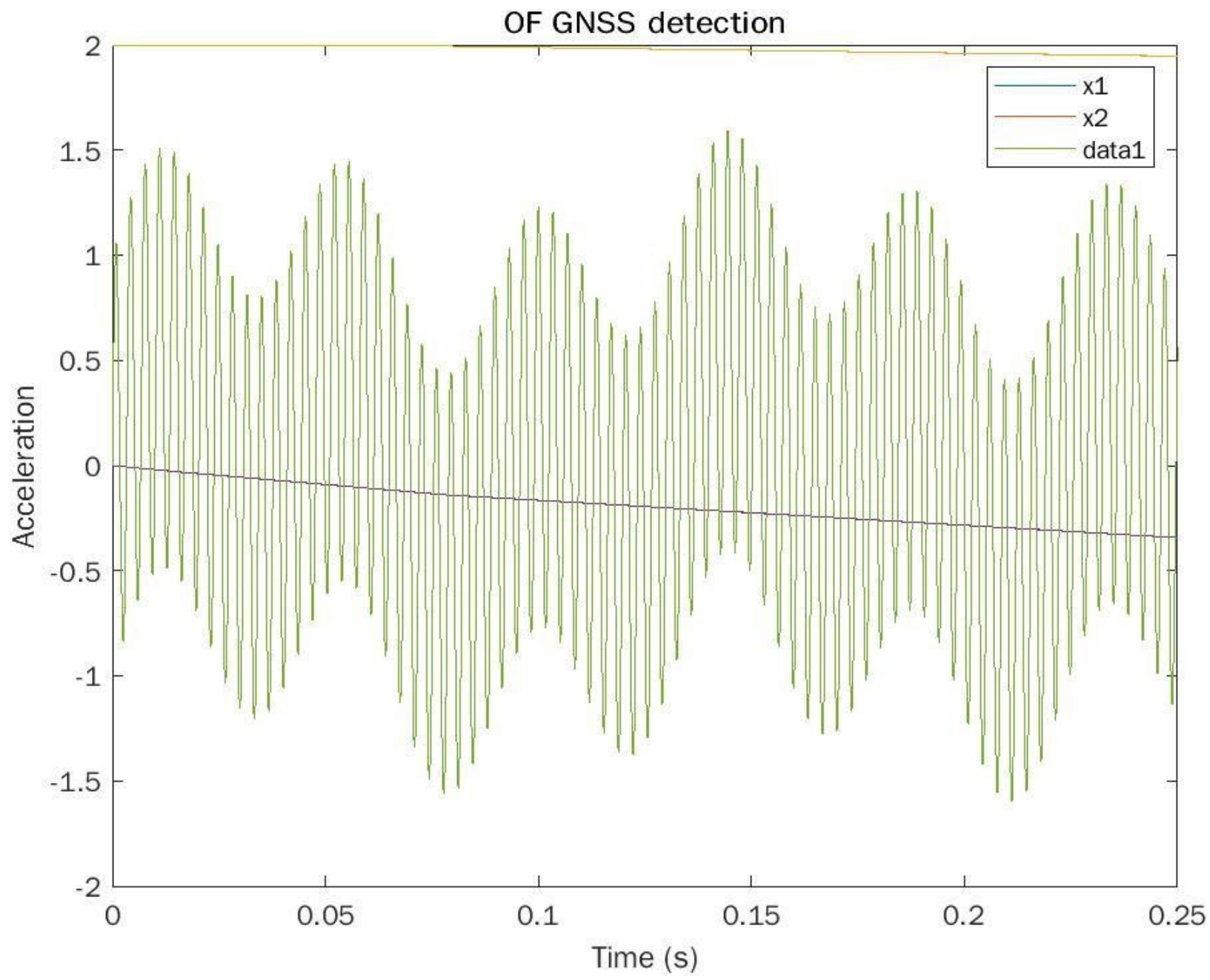

Conclusion:

We have used optical flow in GNSS system for changes in time, position and velocity. Space science is ever increasing with their postulates coming up very rapidly. Though most is based on hypothesis for most visible theories not much proof or concrete data is available for use. We have shown how this can be implemented in a strava similar app for uses in various applications in tracking in the indian subcontinent. Besides I would like to emphasize on our work on optical flow GNSS which is used to determine the accuracy of the vehicles or moving body on the surface of the earth with respect to the satellite. This is a new theory put forth inorder to accelerate the process of global positioning system detection. Optical flow was formally derived from the fundamental theorem of calculus which formally put forward in the paper by Horn and Schunk in determining for optical flow paper. We have in our paper experimentally shown flaring horn shaped antenna, besides simulations obtained on the GPS sensor and on point to point GNSS detection. The optical flow GNSS proves to be very accurate in GPS system design in navigation in the indian subcontinent(NavIC) there is an improvement in our algorithm working as compared to the previous paper [19] where they propose ionospheric models and [20] which proposs new hybrid constellation. We have detected accurate theorem for this purposes. The proposed algorithm is in variant and seen for earlier theorems in global positioning. This theorem can be used in the future in car GPS systems for increased efficiency.

Acknowledgement:I would like to express my gratitude to Indian institute of technology for their support in this project in addition to Chandigarh University in allowing us to conduct this experiment.